\documentclass[letterpaper, 10 pt, conference]{ieeeconf}  % Comment this line out if you need a4paper

\IEEEoverridecommandlockouts                              % This command is only needed if 
\usepackage{graphics} % for pdf, bitmapped graphics files
\usepackage{times} % assumes new font selection scheme installed
\usepackage{amsmath} % assumes amsmath package installed
\usepackage{amssymb}  % assumes amsmath package installed

\usepackage{times}
\usepackage{epsfig}
\usepackage{graphicx}
\usepackage{amsmath}
\usepackage{amssymb}
\usepackage{stfloats}
\usepackage{amssymb}

\usepackage[noend]{algpseudocode}
\usepackage{algorithm}
\usepackage{xspace}

\usepackage{hyperref} 
\usepackage{amsfonts}
\usepackage{booktabs}
\usepackage{siunitx}
\usepackage{multirow}

\DeclareMathAlphabet{\mathpzc}{OT1}{pzc}{m}{it}

\newcommand{\mathkomma}{\quad ,}

\usepackage{color}

\newcommand{\sn}[1]{\textit{SalsaNet }{#1}} 
\newcommand{\sqa}[1]{SqSeg-V1 {#1}}
\newcommand{\sqb}[1]{SqSeg-V2 {#1}}  
\newcommand{\un}[1]{U-Net {#1}}

\def\eg{e.g.\@\xspace}
\def\ie{i.e.\@\xspace}

\title{\LARGE \bf
SalsaNet: Fast Road and Vehicle Segmentation \\ in LiDAR Point Clouds   for Autonomous Driving
}

\author{Eren Erdal Aksoy$^{1,2}$, Saimir Baci$^{2}$, and Selcuk Cavdar$^{2}$% <-this % stops a space
\thanks{The research leading to these results has received funding from the Vinnova FFI project SHARPEN, under grant agreement no. 2018-05001.}
\thanks{$^{1}$Halmstad University, School of Information Technology, Center for Applied Intelligent Systems Research, Halmstad, Sweden} 
\thanks{$^{2}$Volvo Technology AB, Volvo Group Trucks Technology, Vehicle Automation, Gothenburg, Sweden
}%
}

\begin{document}

\maketitle
\thispagestyle{empty}
\pagestyle{empty}

%%%% to do's
%%% update git repo with a readme file
%%% update the repo with the annotated KITTI point clouds
%%% upload the video
%%% add one more confusion matrix 
%-------------------------------------------------------------------------
\begin{abstract}
  
In this paper, we introduce a deep encoder-decoder network, named \textit{SalsaNet}, for   efficient semantic segmentation of 3D LiDAR point clouds.
\sn segments the  road, i.e. drivable free-space, and vehicles in the scene by employing the Bird-Eye-View (BEV) image projection of the point cloud. 
To overcome the lack of annotated point cloud data, in particular for the road segments, we introduce an \textit{auto-labeling} process which transfers automatically generated labels from the camera  to LiDAR.
We also explore the role of image-like projection of LiDAR data in semantic segmentation by comparing BEV  with spherical-front-view projection and show that \sn is \textit{projection-agnostic}.
% and exhibits the highest performance in both point cloud representations.
%i.e. does not suffer due to the distortion and uncommon deformation in the projection.
%
We perform quantitative and qualitative evaluations on the \textit{KITTI dataset}, which demonstrate that the proposed   \sn outperforms other state-of-the-art semantic segmentation networks in terms of  accuracy and computation time. Our code and data are publicly available at \href{https://gitlab.com/aksoyeren/salsanet.git}{https://gitlab.com/aksoyeren/salsanet.git}.
 
\end{abstract}

%-------------------------------------------------------------------------
\section{Introduction}

Semantic segmentation of 3D point cloud data  plays a key role in scene understanding to reach full autonomy in self-driving vehicles.
For instance, estimating the free drivable space together with vehicles in the front can lead to safe maneuver  planning and decision making, which enables autonomous driving to a great extent.

Recently, great progress has been made in deep learning to generate accurate, real-time and robust semantic segments.  
Most of these advanced segmentation approaches heavily rely on camera data \cite{FastSCNN2019,Meyer2019,Frustum2017}.
In contrast to passive camera sensors, 3D LiDARs (Light Detection And Ranging) have wider field of view and provide significantly reliable distance measurement  robust to environmental illumination. Thus, 3D LiDAR scanners have always been an important component in the perception pipeline of autonomous vehicles.

Unlike images, LiDAR point clouds are, however, relatively sparse and contain a vast number of irregular, i.e. unstructured, points. 
In addition, the density of points varies drastically due to  non-uniform sampling of the environment, which makes the intensive point searching and indexing operations relatively expensive. 
Among others, a common attempt to tackle all these challenges is to project point clouds into a 2D image space in order to generate a structured (matrix) form required for the standard convolution process. Existing 2D projections are Bird-Eye-View (BEV) (\ie top view) and Spherical-Front-View (SFV) (\ie panoramic view). 
%Although these projections yield compact representations, SFV  introduces deformation  to the projection of small objects, \eg vehicles in the scene.  
However, to the best of our knowledge, there is still no comprehensive comparative  study showing the contribution of these projection methods in the segmentation process. 
%Although, these projections yield compact representations, they introduce information loss and additional deformation in object size and shape during image discretization. 
%Thus, the trained model with such representations are more biased with the  distortion and deformation in the projection.
%Therefore, the segmentation network should be invariant to the  distortion and deformation in the projection. 

In the context of semantic segmentation of 3D LiDAR data, most of the recent studies employ these   projection methods
to focus on the estimation of either the road itself \cite{Caltagirone2017,Velas18}  or only the obstacles on the road (\eg vehicles) \cite{SqueezesegV01,SqueezesegV02}. All these segments are, however, equally important for the subsequent navigation components (e.g. maneuver  planning) and, thus, need to be jointly processed. 
The main reason of having this decoupled treatment in the literature is the lack of large annotated  point cloud    data, in particular, for the road segments.

In this paper, we study the joint segmentation of  the road, i.e. drivable free-space, and vehicles using  3D LiDAR point clouds only. 
We propose a novel ``SemAntic Lidar data SegmentAtion Network", \ie \textit{SalsaNet}, which has an encoder-decoder architecture where the encoder part contains consecutive ResNet blocks  \cite{resnet2016}. 
Decoder part rather   upsamples features and combines them with the corresponding counterparts from the early residual blocks via skip connections. Final output of the decoder is then sent to the   pixel-wise classification layer to return semantic segments.  

We validate our network's performance on the KITTI dataset \cite{KittiDataset} which provides 3D bounding boxes for vehicles  and a relatively small number  of annotated road images ($\approx$300 samples). Inspired from  \cite{Piewak18, Wang19}, we propose an \textit{auto-labeling} process to automatically label $\approx$11K point clouds in the KITTI dataset.
For this purpose, we employ the state-of-the-art methods \cite{Teichmann2018} and \cite{maskrcnn} to respectively segment road and vehicles in camera images. These segments are then mapped from camera space to LiDAR to automatically generate annotated point clouds. 

The input of \sn is the BEV rasterized image format of the point cloud  where each image channel stores a unique statistical cue (\eg average depth and intensity values).   
To further analyze the role of the point cloud projection in the network performance, we separately train \sn with the SFV representation and provide a comprehensive comparison with the BEV counterpart.  

Quantitative and qualitative experiments on the KITTI dataset  show that the proposed \sn is \textit{projection-agnostic}, \ie exhibiting high performance in both projection methods  and significantly outperforms other  state-of-the-art semantic segmentation approaches \cite{SqueezesegV01,SqueezesegV02,Unet}  in terms of pixel-wise segmentation accuracy while requiring much less computation time.

Our contribution is manifold:
\begin{itemize}
  \item We introduce an   encoder-decoder architecture to semantically segment  road and vehicle points in real-time using 3D LiDAR data only.
  \item To automatically annotate a large set of 3D LiDAR point clouds, we transfer labels across different sensor modalities, e.g. from camera images to LiDAR.
  \item We study two   commonly used point cloud projection methods   and compare their effects in the semantic segmentation process in terms of accuracy and speed.
  \item We provide a thorough quantitative comparison of the proposed \sn on the KITTI dataset with different state-of-the-art semantic segmentation networks. 
  \item We also release our source code and annotated point clouds to encourage further research.
\end{itemize}

%-------------------------------------------------------------------------
\section{Related Work}
 
There exist multiple prior methods exploring the semantic segmentation of 3D LiDAR point cloud data. Those methods are  basically classified under two categories. The first one involves conventional heuristic approaches such as model fitting by employing iterative approaches \cite{Zermas17} or histogram computation after projecting   LiDAR point clouds to 2D space \cite{Chen17}. In contrast, the second category investigates advanced deep learning approaches \cite{SqueezesegV01,SqueezesegV02,PointSeg18,Pixor2019} which achieved a quantum jump in performance during the last decade. These approaches in the latter class differ from each other not only in terms of  network architecture but also in the way the LiDAR data  is represented before being fed to the network. 
 
Regarding the network architecture, high performance segmentation methods particularly use fully convolutional networks \cite{fcn2016}, encoder-decoder structures \cite{Zhang2018}, or multi-branch models \cite{FastSCNN2019}.
The main difference between these architectures is the way of encoding  features at various depths and later incorporating them for recovering the spatial information.
We, in this study,  follow the encoder-decoder structure due to the  high  performance observed in
the   recent  state-of-the-art point cloud segmentation studies  \cite{SqueezesegV01,SqueezesegV02,Unet}.

In the context of 3D LiDAR point cloud representation, there exist three mainstream methods: voxel creation \cite{VoxelNet18,Zhang2018}, point-wise operation \cite{PointNet2016}, and image projection \cite{SqueezesegV01,SqueezesegV02,PointSeg18}. Voxel representation  transforms a point cloud into a high-dimensional volumetric form, i.e. 3D voxel grid \cite{VoxelNet18,Zhang2018}. 
%
%Due to sparsity in point clouds, the voxel grid, however, becomes sparse which leads to redundant computations. 
Due to sparsity in point clouds, the voxel grid, however, may have empty voxels which leads to redundant computations. 
Point-wise methods \cite{PointNet2016} process  points  directly without converting them into any other form. The main drawback here is the processing capacity which cannot efficiently handle large LiDAR point sets unless fusing them with additional cues from other sensory data, such as camera images as shown in \cite{Frustum2017}.
To handle the sparsity in  LiDAR point clouds, various  image space projections, such as  Bird-Eye-View (BEV) (\ie top view) \cite{Caltagirone2017,rt3d2018,ComplexYolo} and Spherical-Front-View (SFV) (\ie panoramic view) \cite{SqueezesegV01,SqueezesegV02,PointSeg18}, have been introduced. In contrast to voxel and point-wise approaches, the 2D projection is more compact, dense and, thus, amenable to real-time computation.
In this study, our network relies on BEV since the SFV projection  introduces   distortion and uncommon deformation which has significant effects on relatively small objects such as vehicles.

Closest to our work are the recent studies in \cite{SqueezesegV01,SqueezesegV02}. Those approaches, however, segment only the road objects such as vehicles, pedestrians and cyclists using the SFV projection. In contrast, we only focus on road and vehicle segments in the poincloud. We exclude pedestrians and cyclists since their total number of instances in the entire KITTI dataset is significantly low which naturally yields poor learning performance as already shown in \cite{SqueezesegV01,SqueezesegV02}. 
 
There are also numerous   networks addressing the road segmentation \cite{Caltagirone2017,Velas18}. These approaches, however, omit road objects (\eg vehicles) and rely on limited annotated data ($\approx$300 samples) in the KITTI road dataset. 
To generate more training data, we automatically label road and vehicle segments in the entire KITTI point clouds ($\approx$11K samples) by transferring the camera image annotations derived by state-of-the-art segmentation methods  \cite{Teichmann2018}, \cite{maskrcnn}. Note that a similar auto-labeling process has already been proposed in \cite{Piewak18, Wang19} and a new large labeled 3D road scene point cloud dataset has been introduced in \cite{Zhang2018}, however, the final labeled data has still not been released in any of these works for public use, which is not the case for our work.

%There are also alternative multi-modal methods that fuse LiDAR data with camera images \cite{MTMS2019,Meyer2019,Frustum2017}. Although these approaches show high performance, they require an additional time synchronized and calibrated  camera and also have to tackle with any sensor failure mode. In this work, we therefore, focus on semantic segmentation of LiDAR-only data stream.

%-------------------------------------------------------------------------
\section{Method}
In this section, we give a detailed description of our proposed method starting with the automatic labeling of 3D  point cloud data. We then continue with the point cloud representation, network architecture and training details.

\subsection{Data Labeling}
\label{sec:labeling}
\begin{figure*}[!t]
    \centering
    \includegraphics[scale=0.365]{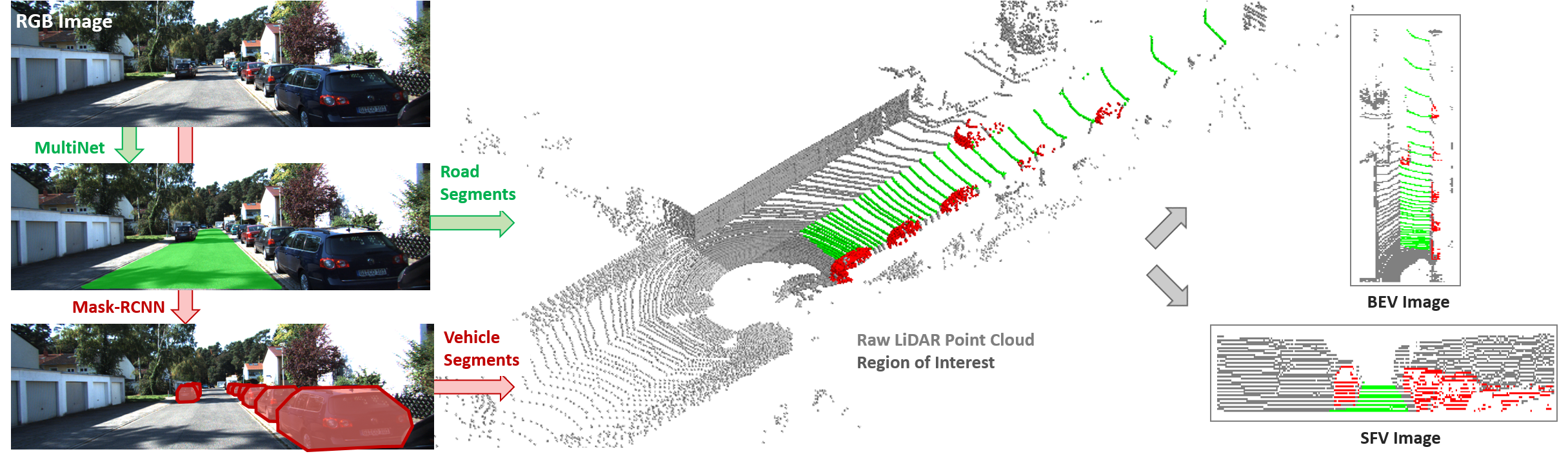}
    \caption{Automatic point cloud labelling to generate network inputs in BEV and SFV formats.}
    \label{fig:labeling}
\end{figure*}

Lack of large annotated  point cloud data introduces a challenge in the network training and evaluation. Applying crowdsourced manual data labeling is cumbersome due to a huge number of points in each individual scene cloud. Inspired from  \cite{Piewak18, Wang19}, we, instead, propose an \textit{auto-labeling} process illustrated in Fig.~\ref{fig:labeling} to automatically label 3D LiDAR point clouds.  

Since the image-based detection and segmentation methods are relatively more mature than LiDAR-based solutions, we benefit from  this stream of work to annotate 3D LiDAR point clouds. In this respect, to extract the road pixels, we use an off-the-shelf neural network MultiNet \cite{Teichmann2018}  dedicated to the road segmentation in camera images. We here note that the reason of using MultiNet is beacuse it is open-source and is already trained on   the KITTI road detection benchmark  \cite{KittiDataset}.   
To extract the vehicle points in the cloud, we  employ Mask R-CNN \cite{maskrcnn} to segment labels in the camera image.
Note that in case of having  bounding box object annotations, as in the KITTI object detection benchmark \cite{KittiDataset}, we directly label those points inside the 3D bounding box as vehicle segments.
Finally, with the aid of the camera-LiDAR calibration, both road and vehicles segments are  transferred from   image space to the point cloud  as shown in  Fig.~\ref{fig:labeling}.

\subsection{Point Cloud Representation}
\label{sec:projection}

Given an unstructured 3D LiDAR point cloud, we investigate the 2D grid-based image representation since it is more compact  and yields real-time inference. As 3D points lies on a grid in the projected form, it also allows standard convolutions. We here note that   reducing the dimension from 3 to 2 does not yield information loss in the point cloud since the height information is still kept as an additional channel in the projected image.
Therefore, in this work, we exploit two common projections: Bird-Eye-View (BEV) and Spherical-Front-View (SFV) representations.

\subsubsection{Bird-Eye-View (BEV)} 
\label{sec:bev}

We initially define a region-of-interest in the raw point cloud, covering a large area  in front of the vehicle which is  $50m$ long ($ x\in[0,50]$) and  $18m$ wide ($ y\in[-6,12]$) (see the dark gray points in Fig.~\ref{fig:labeling}).
All the 3D points in this region-of-interest are then projected and discetized in to a 2D grid map with the size of $256\times64$. 
The grid map corresponds to the $x-y$ plane of the LiDAR and forms the BEV, \ie top view  projection of the point cloud.
We  set the grid cell sizes to $0.2$ and $0.3$ in $x-$ and $y-$axes, respectively. 
A sample BEV image   is depicted in Fig.~\ref{fig:labeling}.

Similar to the work in \cite{Caltagirone2017}, in each grid cell, we compute the mean and maximum elevation, average reflectivity (\ie intensity) value, and number of projected points. Each of these four statistical information is encoded as one unique image channel, forming a 4D BEV image to be further used as the network input. Note that we also normalize each image channel to be within $[0,1]$. 
Compared to \cite{Caltagirone2017}, we avoid using the minimum and standard deviation values of the height as additional   features	 since our experiments  showed that there is no significant contribution coming from those channels.

\subsubsection{Spherical-Front-View (SFV)}
\label{sec:sfv}

Following the work of \cite{SqueezesegV01}, we also project the 3D point cloud onto a sphere to generate dense grid image representation in a rather panoramic view. 

In this projection, each point is represented by two angles ($\theta,\phi$) and an intensity ($i$) value. In the 2D spherical grid image, each point is mapped to the coordinates ($u,v$), where $u=\lfloor \theta/\Delta\theta\rfloor$ and  $v=\lfloor\phi/\Delta\phi\rfloor$.  
Here, $\theta$ and $\phi$ are azimuth and zenith angles computed from point coordinates ($x,y,z$) as $\theta = \arcsin(z/\sqrt{x^2+y^2+z^2}) $  and $\phi = \arcsin(y/\sqrt{x^2+y^2})$, whereas $\Delta\theta$ and $\Delta\phi$ define the discretization resolutions. 

For the projection, we   mark the front-view area of $90^\circ$ as a region-of-interest.  In each grid cell, we separately store 3D Cartesian coordinates ($x,y,z$), the intensity value ($i$) and range $r=\sqrt{x^2+y^2+z^2}$. As in \cite{SqueezesegV02}, we also keep a binary mask indicating the occupancy of the grid cell. These six channels form the final   image which has the resolution of $64\times512$. A sample SFV image   is depicted in Fig.~\ref{fig:labeling}.

Although SFV returns more dense representation compared to BEV, SFV has certain distortion and deformation effects on  small objects, \eg vehicles. 
It is also more likely that objects in SFV tend to occlude each other. We, therefore, employ BEV representation as the main input to our network. We, however, compare these two projections in terms of their contribution to the segmentation accuracy.

\subsection{Network Architecture}

\begin{figure*}[!t]
    \centering
    \includegraphics[scale=0.69]{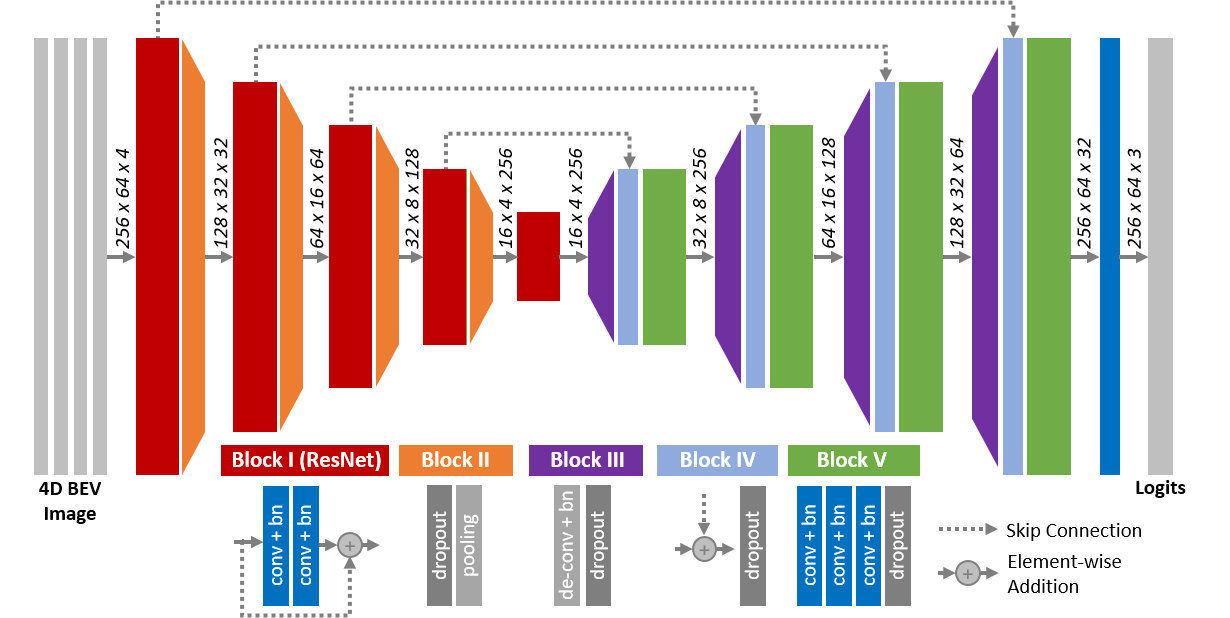}
    \caption{Architecture of the proposed \textit{SalsaNet}.  Encoder part involves a series of ResNet blocks. Decoder part upsamples feature maps and combines them with the corresponding  early residual block outputs using skip connections. Each convolution layer in Blocks I-V   is coupled with   a leaky-ReLU activation layer and a batch normalization (bn) layer.}
    \label{fig:salsanet}
\end{figure*}

The   architecture of the proposed \sn is depicted in Fig.~\ref{fig:salsanet}.
The input to \sn is a $256\times64\times4$ BEV projection of the point cloud as described in Sec.\ref{sec:bev}.

\sn has an encoder-decoder structure where the encoder part contains a series of ResNet blocks  \cite{resnet2016} (Block I in Fig.~\ref{fig:salsanet}). Each ResNet block, except the very last one, is followed by dropout and pooling layers (Block II in Fig.~\ref{fig:salsanet}). 
We employ   \textit{max-pooling}   with kernel size of $2$ to downsample feature maps in both width and height. Thus, on the encoder side, the total downsampling factor goes up to 16. 
Each convolutional layer has kernel size of $3$, unless otherwise stated.
The number of feature channels are respectively $32, 64, 128, 256,$ and $256$.
Decoder network has a sequence of deconvolution layers, \ie transposed convolutions (Blocks III in Fig.~\ref{fig:salsanet}), to upsample feature maps, each of which is then element-wise added to the corresponding lower-level (bottom-up)  feature maps of the same size transferred via   skip connections (Blocks IV in Fig.~\ref{fig:salsanet}).
After each feature addition in the decoder, a stack of convolutional layers (Blocks V in Fig.~\ref{fig:salsanet}) are introduced to capture more precise spatial cues to be further propagated to the higher layers. 
The next layer applies $1\times1$ convolution to have $3$ channels which corresponds to the total number of semantic classes (\ie road, vehicle, and background). 
Finally, the  output feature map  is  fed to a soft-max classifier to obtain pixel-wise classification.

Each convolution layer in Blocks I-V (see  Fig.~\ref{fig:salsanet}) is coupled with   a leaky-ReLU activation layer.
We further  applied batch normalization \cite{ioffe2015batch} after each convolution in order to help converging to the optimal solution by solving the internal covariate shift. 
We here  emphasize  that dropout needs to be placed right after batch normalization. As shown in \cite{li2018understanding}, an early application of dropout can otherwise lead to a shift in the weight distribution and thus minimize the effect of batch normalization during training.

\subsection{Class-Balanced Loss Function}
\label{sec:loss}

%  See section 3.4 in this paper \cite{Zhang2018} to describe the weighted  loss function 
% https://pdfs.semanticscholar.org/00a4/e3f469ff6e1e90d87e4c6aca7210266f6f84.pdf

Publicly available datasets mostly have an extreme imbalance between different classes. For instance, in the autonomous driving scenarios, vehicles appear less in the scene compared to road and background. Such an imbalance between classes yields the network to be more biased to the classes that have more samples in training and thus results in relatively poor segmentation results. 

To value more the under-represented classes, we update the softmax cross-entropy loss with the smoothed frequency of each class. Our class-balanced loss function is now weighted with the inverse square root of class frequency, defined as  
 
\begin{equation} \label{lossequation}
    \mathcal{L}(y,\hat{y}) = -\sum_{i} \alpha_{i}p(y_{i})log(p(\hat{y}_{i}))  \mathkomma
\end{equation}

\begin{equation}
        \alpha_{i} = 1/\sqrt{f_{i}}    \mathkomma
\end{equation}

where $y_i$ and $\hat{y}_{i}$ are the true and predicted labels and $f_{i}$ is the frequency, i.e. the number of points, of the $i^{th}$ class. This helps the network strengthen each pixel of classes that appear less in the dataset.

\subsection{Optimizer And Regularization}

To train \sn we employ the Adam optimizer \cite{AdamOpt} with the initial learning rate of $0.01$ which is decayed by $0.1$ after every $20K$ iterations. The dropout  probability and batch size are  set to $0.5$ and $32$, respectively. We run the training for $500$ epochs.

To increase the amount of training data, we also augment the network input data by flipping horizontally, adding random pixel noise with probability of $0.5$, and randomly rotating about the $z-$axis  in the range of $[-5^{\circ},5^{\circ}]$.

%-------------------------------------------------------------------------
\section{Experiments}
  
To show both the strengths and weaknesses of our model, we evaluate the performance of \sn  and compare with the other state-of-the-art semantic segmentation methods on the challenging KITTI dataset  \cite{KittiDataset} which provides 3D LiDAR scans. We first employ the \textit{auto-labeling} process described in Sec.~\ref{sec:labeling} to acquire point-wise annotations. In total, we generate $10,848$ point clouds where each point is labeled to one of $3$ classes, \ie road, vehicle, or background.
We then follow  exactly the same protocol  in \cite{SqueezesegV01} and divide the KITTI dataset into training and test splits with $8,057$ and $2,791$ point clouds.  
We implement our model in TensorFlow and release the code and labeled point clouds for public use\footnote{\href{https://gitlab.com/aksoyeren/salsanet.git}{https://gitlab.com/aksoyeren/salsanet.git}}.

\subsection{Evaluation Metric}
The performance of our model is measured on class-level segmentation tasks by comparing each predicted point label with the corresponding ground truth annotation.
As the primary evaluation metrics, we report precision ($\mathrm{P}$), recall ($\mathrm{R}$), and intersection-over-union ($\mathrm{IoU}$) results for each individual class as
\[
\mathrm{P}_i=\frac{|\mathcal{P}_i\cap\mathcal{G}_i|}{|\mathcal{P}_i|},~
\mathrm{R}_i=\frac{|\mathcal{P}_i\cap\mathcal{G}_i|}{|\mathcal{G}_i|},~
\mathrm{IoU}_i=\frac{|\mathcal{P}_i\cap\mathcal{G}_i|}{|\mathcal{P}_i\cup\mathcal{G}_i|}, ~
\]
where $\mathcal{P}_i$ is the predicted point set of class $\textit{i}$ and $\mathcal{G}_i$ denotes the corresponding ground truth set, whereas $|.|$ returns the total number of points in a set. In addition, we report the average IoU score over all the three classes.

%-------------------------------------------------------------------------
\newcommand{\ra}[1]{\renewcommand{\arraystretch}{#1}}
\begin{table*}\centering
\scalebox{0.8}{
\ra{1.3}
\begin{tabular}{@{}cclcccccccccccccc@{}}\toprule
&&& \multicolumn{3}{c}{Background} & \phantom{abc}& \multicolumn{3}{c}{Road} & \phantom{abc} & \multicolumn{3}{c}{Vehicle} & \phantom{abc} & Average &\\
\cmidrule{4-6} \cmidrule{8-10} \cmidrule{12-14}  \cmidrule{16-16}
                            &&& Precision & Recall & IoU    && Precision & Recall & IoU    && Precision & Recall & IoU    && IoU &  \\ \midrule
%%%%%%%%%%%%%%%%%% bird eye view %%%%%%%%%%%%%%%%%%%%%%%%%%%%%%%%%%%%%%%%%                          
&\multirow{4}{*}{\rotatebox[origin=c]{90}{\textit{BEV}}}   
& SqSeg-V1~\cite{SqueezesegV01}  & 99.53  & 97.87  & 97.42  && 72.42  & 89.71 & 66.86 && 46.66 & 93.21  & 45.13  && 69.80  &\\
&& SqSeg-V2~\cite{SqueezesegV02} & 99.39  & 98.59  & 97.99  && 77.26  & 87.33 & 69.47 && 66.28 & 90.42  & 61.93  && 76.46  &\\
&& U-Net~\cite{Unet}             & 99.54  & 98.47  & 98.03  && 76.08  & 90.84 & 70.65 && 67.23 & 90.54  & 62.80  && 77.27  &\\
&& Ours                         & 99.46  & 98.71  & \textbf{98.19}  && 78.24 & 89.39   & \textbf{71.61} && 75.13  & 89.74   & \textbf{69.19} && \textbf{79.74} &\\
\bottomrule
%%%%%%%%%%%%%%%%%% front view %%%%%%%%%%%%%%%%%%%%%%%%%%%%%%%%%%%%%%%%%   
&\multirow{4}{*}{\rotatebox[origin=c]{90}{\textit{SFV}}} 
& SqSeg-V1~\cite{SqueezesegV01}  & 97.09  & 95.02  & 92.39  && 79.72  & 83.39  & 68.79 && 70.70  & 91.50  & 66.34  && 75.84 &\\
&& SqSeg-V2~\cite{SqueezesegV02} & 97.43  & 95.73  & 93.37  && 80.98  & 86.22  & 71.70 && 77.25  & 89.48  & 70.82  && 78.63 &\\
&& U-Net~\cite{Unet}             & 97.84  & 94.93  & 92.98  && 78.14  & 88.62  & 71.00 && 74.65  & 90.57  & 69.26  && 77.84 &\\
&& Ours                         & 97.81  & 95.76  & \textbf{93.75}  && 81.62  & 88.38  & \textbf{73.72} && 77.03  & 90.79  & \textbf{71.44}  && \textbf{79.71} &\\
\bottomrule
\end{tabular}
}
\caption{Quantitative comparison on KITTI's test set. Scores are given in percentage ($\%$).} 
\label{tab:quanresults}
\end{table*}
%-------------------------------------------------------------------------

\subsection{Quantitative Results}
 
We compare the performance of \textit{SalsaNet} with the other state-of-the-art networks: SqueezeSeg (SqSeg-V1) \cite{SqueezesegV01} and SqueezeSegV2 (SqSeg-V2) \cite{SqueezesegV02}. 
We particularly focus on these specialized networks because they are implemented only for the semantic segmentation task,  solely rely on 3D LiDAR point clouds, and also provide open-source implementations.  
We train the networks \sqa   and  \sqb with the same configuration parameters provided in \cite{SqueezesegV01} and \cite{SqueezesegV02}. To obtain the highest score, we, however, alter  the learning rate (set to $0.001$ for SqSeg-V1) and also apply the same data augmentation protocol used for the training of \textit{SalsaNet}.
As an additional baseline method, we   implement a vanilla \un model \cite{Unet} since it is structurally similar to \textit{SalsaNet}. 
For a fair comparison, we train \un with exactly the same parameters (\eg learning rate, batch size, etc.) and strategy (\eg loss function, data augmentation, etc.) used for the training of \textit{SalsaNet}.
We run our experiments for both BEV and SFV  projections to study the effect of LiDAR point cloud projection on semantic segmentation. 
Obtained quantitative results are reported in Table~\ref{tab:quanresults}.

In all cases, our proposed model \sn considerably outperforms the others by leading to the highest IoU scores.
In BEV, \sn particularly performs well on vehicles which are relatively small objects (\eg compared to the road). In other methods, the highest IoU score for vehicles is $6.3\%$ less than that of \textit{SalsaNet}, which clearly indicates that these methods have difficulties extracting the local features in BEV projection. When it comes to the SFV projection, this margin between the vehicle IoU scores  shrinks to $0.6\%$ although \sn still performs the best. It is because SFV has more compact form: small objects like vehicles become bigger while relatively bigger objects (such as background) occupy less portion of the SFV image (see Fig.~\ref{fig:labeling}). This finding indicates that \sn is \textit{projection-agnostic} as it can capture the local features, \ie performs equivalently well, in both projection methods.

\begin{figure}[!b]
    \centering
    \includegraphics[scale=0.27]{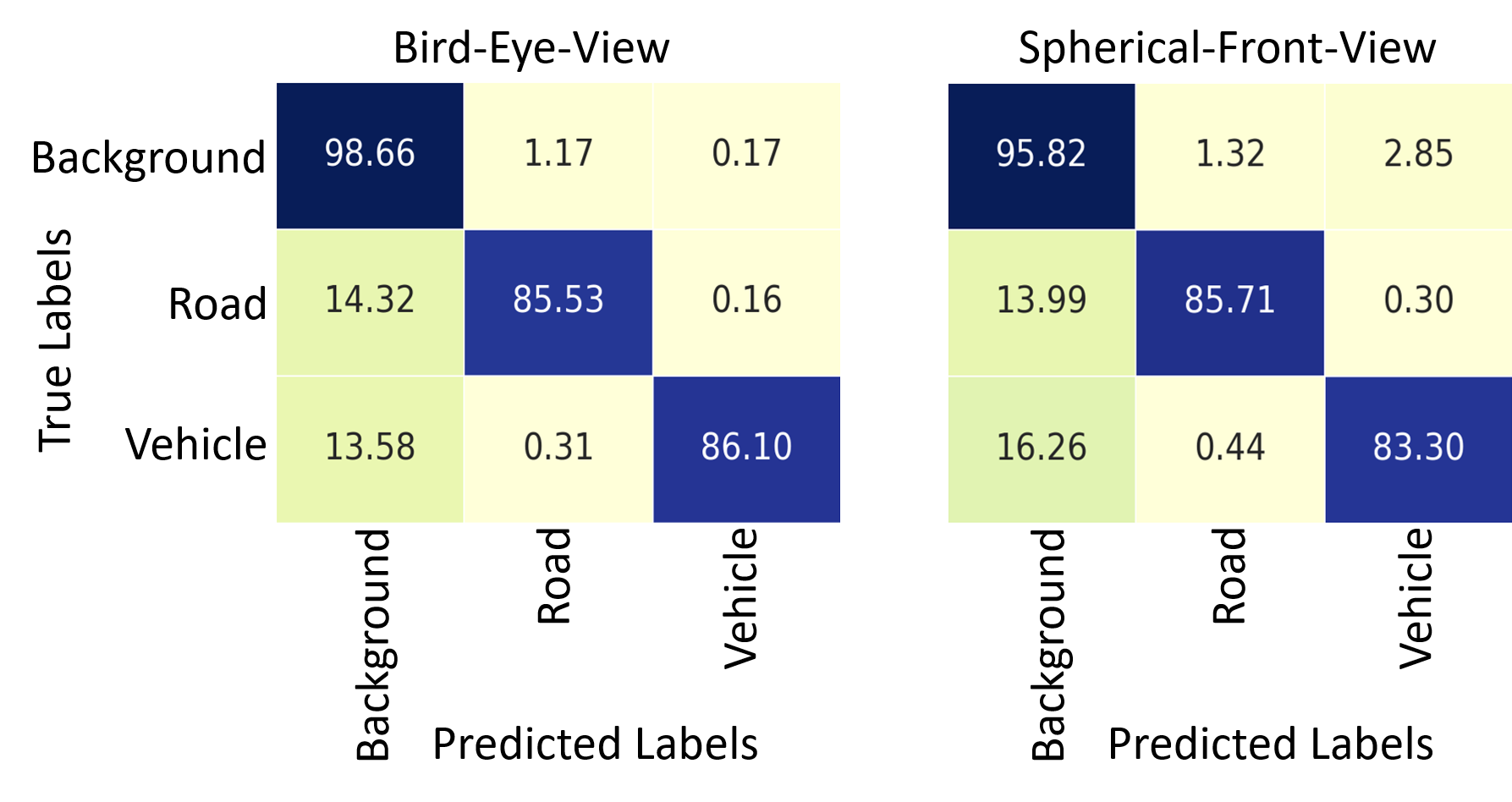}
    \caption{The confusion matrices for  our model \textit{SalsaNet} using BEV and SFV projections on  KITTI's test split. }
    \label{fig:confmat}
\end{figure}

Furthermore, Fig.~\ref{fig:confmat} depicts the final confusion matrices for \sn using both BEV and SFV projections. This figure clearly presents that there is no major confusion between the classes. A small number of vehicle and road points are labeled as background but not mixed with each other. We believe that points on the road and vehicle borders in both image representations cause this minor mislabeling which can easily be overcome with more precise annotation of the training data.

\begin{figure*}[!t]
    \centering
    \includegraphics[scale=0.47]{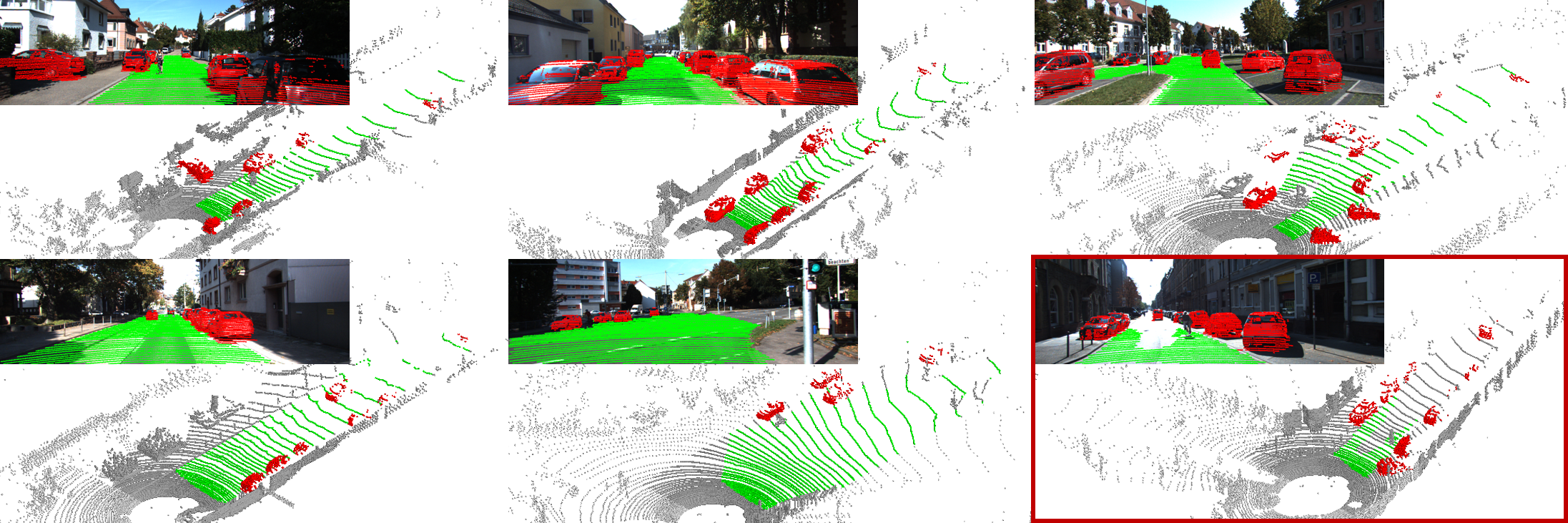}
    \caption{Sample qualitative results showing successes and failures of our proposed method using BEV [best view in color]. Note that the corresponding camera images on the top left are only  for visualization purposes and have not been used in the training. The dark- and light-gray points in the point cloud represent points that are inside and outside the bird-eye-view region, respectively. The green and red points indicate road and vehicle segments.}
    \label{fig:segresults}
\end{figure*}

%-------------------------------------------------------------------------
\subsection{Qualitative Results}

For the qualitative evaluation, Fig.~\ref{fig:segresults} shows some sample semantic segmentation results generated by \textit{SalsaNet} using BEV. In this figure, only for visualization purpose,  segmented road and vehicle points are also projected back to the respective camera image. We, here, emphasize that these camera images have not been used for training of \textit{SalsaNet}.

As depicted in Fig.~\ref{fig:segresults}, \sn can, to a great extent, distinguish road, vehicle, and background points. All other excluded classes, \eg cyclists on the road as shown in the first, fifth, and sixth frames in Fig.~\ref{fig:segresults}, are correctly segmented as background.
We also illustrate a failure case in the last frame of Fig.~\ref{fig:segresults}. 
In this case, the road segment is incomplete. 
It is because the ground truth of the road segment only relies on the output of MultiNet \cite{Teichmann2018} (see Sec.~\ref{sec:labeling}) which, however, returns missing segments due to overexposure of the road from strong sunlight  (see the camera image in the red frame in Fig.~\ref{fig:segresults}).  As a potential solution, we are planning to employ a more accurate road segmentation network  for camera images   to increase the labeling quality of point clouds in the training dataset.

In the supplementary video\footnote{\href{https://youtu.be/grKnW-uGIys}{https://youtu.be/grKnW-uGIys}}, we provide more qualitative results on various   KITTI scenarios.

%-------------------------------------------------------------------------
\begin{table}[!b]\centering
\scalebox{0.69}{
\ra{1.3}
\begin{tabular}{@{}cccccccccccccc@{}}\toprule
&& \multicolumn{4}{c}{Channels} & \multicolumn{3}{c}{Loss} &   \multicolumn{4}{c}{IoU} \\
\cmidrule{3-6} \cmidrule{8-8} \cmidrule{10-13} 
&& Mean & Max & Ref & Dens  &&Weight&& Background & Road  & Vehicle & Average  \\ \midrule
%%%%%%%%%%%%%%%%%% bird eye view %%%%%%%%%%%%%%%%%%%%%%%%%%%%%%%%%%%%%%%%%                          
&\multirow{6}{*}{\rotatebox[origin=c]{90}{\textit{Bird-Eye-View}}}   
&\checkmark &\checkmark &\checkmark &\checkmark &&\checkmark && 98.19 &\textbf{71.61} &\textbf{69.19} &\textbf{79.74}\\
&& \checkmark & \checkmark & \checkmark & \checkmark && -   && \textbf{98.23}       & 71.52        &  66.70        &  78.91 \\
&& - & \checkmark & \checkmark & \checkmark && \checkmark           && 98.13         & 71.30        &  66.70        &  78.81 \\
&& \checkmark & - & \checkmark & \checkmark && \checkmark           && 98.20         & 71.46        &  67.98        &  79.33 \\
&& \checkmark & \checkmark & - & \checkmark && \checkmark           && 98.13         & 71.02        &  62.46        &  77.30 \\
&& \checkmark & \checkmark & \checkmark & - && \checkmark           && 98.13         & 71.04        &  67.42        &  78.95 \\
\bottomrule
\end{tabular}
}
\caption{Ablative analysis for bird-eye-view. Channels mean, max, ref, and dens stand for the mean and maximum elevation, reflectivity, and number of projected points in order.} 
\label{tab:ablationBEV}
\end{table}
%-------------------------------------------------------------------------

%-------------------------------------------------------------------------
\subsection{Ablation Study}

In this ablative analysis, we investigate the individual contribution of each BEV and SFV image channel in the final performance of \textit{SalsaNet}. We also diagnose the effect of using weighted loss introduced in Eg.~\ref{lossequation} in section~\ref{sec:loss}. 

Table~\ref{tab:ablationBEV} shows the obtained results for BEV. 
The first impression that this table conveys is that excluding  weights in the loss function leads to certain accuracy drops. In particular, the vehicle IoU score decreases by  $2.5\%$, whereas the background IoU score slightly increases (see the second row in Table~\ref{tab:ablationBEV}). This apparently means that the network tends to mislabel road and vehicles points since they are under-represented in the training data. 
IoU scores between the third and sixth rows in the same table show that features embedded in BEV image channels have almost equal contributions to the segmentation accuracy. This clearly indicates that the BEV projection that \sn employs as an input does not have any redundant information encoding.

%-------------------------------------------------------------------------

\begin{table}[!b]\centering
\scalebox{0.68}{
\ra{1.3}
\begin{tabular}{@{}cccccccccccccccc@{}}\toprule
&& \multicolumn{6}{c}{Channels} & \multicolumn{3}{c}{Loss} &   \multicolumn{4}{c}{IoU} \\
\cmidrule{3-8} \cmidrule{10-10} \cmidrule{12-15} 
&& X & Y & Z & I & R & M &&Weight&& Background & Road  & Vehicle  & Average  \\ \midrule
%%%%%%%%%%%%%%%%%% front view %%%%%%%%%%%%%%%%%%%%%%%%%%%%%%%%%%%%%%%%%                          
&\multirow{8}{*}{\rotatebox[origin=c]{90}{\textit{Spherical-Front-View}}}   
& \checkmark &\checkmark &\checkmark & \checkmark & \checkmark & \checkmark && \checkmark && 93.75& 73.72 & \textbf{71.44} & 79.71\\
&&\checkmark &\checkmark &\checkmark & \checkmark & \checkmark & \checkmark && -    && 93.90  & 73.80 &  70.30 &  79.43 \\
&& - & \checkmark & \checkmark & \checkmark & \checkmark & \checkmark && \checkmark && 93.76  & 73.97 &  71.34 &  79.75 \\
&& \checkmark & - & \checkmark & \checkmark & \checkmark & \checkmark && \checkmark && \textbf{94.00}  & 74.61 &  71.21 &  \textbf{80.05} \\
&& \checkmark & \checkmark & - & \checkmark & \checkmark & \checkmark && \checkmark && 93.90  & \textbf{74.72} &  69.77 &  79.50 \\
&& \checkmark & \checkmark & \checkmark & - & \checkmark & \checkmark && \checkmark && 93.07  & 72.87 &  66.06 &  77.41 \\
&& \checkmark & \checkmark & \checkmark & \checkmark & - & \checkmark && \checkmark && 93.19  & 73.73 &  64.57 &  77.30 \\
&& \checkmark & \checkmark & \checkmark & \checkmark & \checkmark & - && \checkmark && 92.46  & 73.29 &  59.53 &  75.20 \\
\bottomrule
\end{tabular}
}
\caption{Ablative analysis for spherical-front-view. Channels x, y, z, i, r, and m stand for the cartesian coordinates ($x,y,z$), intensity, range, and occupancy mask, in order.} 
\label{tab:ablationSFV}
\end{table}

%-------------------------------------------------------------------------

Table~\ref{tab:ablationSFV} shows the  results obtained  for SFV. We, again, observe the very same effect on IoU scores when the applied loss weights are omitted. The most interesting findings, however, emerge when we start measuring the contribution of SFV image channels. As the last six rows in Table~\ref{tab:ablationSFV} indicate, SFV image channels have rather inconsistent effects on the segmentation accuracy. For instance, while adding the mask channel increases the average accuracy by $4.5\%$, the first two channels that keep  x-y coordinates lead to drop in overall average accuracy by $0.04\%$ and $0.34\%$, respectively. This is a clear evidence that the SFV projection introduced in \cite{SqueezesegV01} and \cite{SqueezesegV02} contains redundant information that mislead the feature learning in networks.  

Given these findings and also the arguments regarding the   deformation in SFV as stated in section~\ref{sec:sfv}, we conclude that the BEV projection is more appropriate point cloud representation. Thus, \sn relies on   BEV.

%-------------------------------------------------------------------------
\begin{table}[!t]\centering
\scalebox{0.8}{
\ra{1.3}
\begin{tabular}{@{}cclccc@{}}\toprule
                            &&& Mean (msec) & Std (msec)  & Speed (fps)  \\ \midrule
%%%%%%%%%%%%%%%%%% bird eye view %%%%%%%%%%%%%%%%%%%%%%%%%%%%%%%%%%%%%%%%%                          
&\multirow{4}{*}{\rotatebox[origin=c]{90}{\textit{BEV}}}   
& SqSeg-V1~\cite{SqueezesegV01}  & ~6.77  & ~0.31 &  148 Hz \\
&& SqSeg-V2~\cite{SqueezesegV02} & 10.24 & ~0.31 & ~98 Hz\\
&& U-Net~\cite{Unet}             & \textbf{~5.31}  & ~0.21 & \textbf{188} Hz \\
&& Ours                         & ~6.26  & ~0.08 & 160 Hz\\
\bottomrule
%%%%%%%%%%%%%%%%%% front view %%%%%%%%%%%%%%%%%%%%%%%%%%%%%%%%%%%%%%%%%   
&\multirow{4}{*}{\rotatebox[origin=c]{90}{\textit{SFV}}} 
& SqSeg-V1~\cite{SqueezesegV01}  & ~6.92  & ~0.21 & 144 Hz \\
&& SqSeg-V2~\cite{SqueezesegV02} & 10.36 & ~0.41 & ~96 Hz \\
&& U-Net~\cite{Unet}             & \textbf{~5.45}  & ~0.21 & \textbf{183} Hz  \\
&& Ours                         & ~6.41  & ~0.13  & 156 Hz\\
\bottomrule
\end{tabular}
}
\caption{Runtime performance on KITTI dataset \cite{KittiDataset}} 
\label{tab:runtime}
\end{table}
%-------------------------------------------------------------------------
\subsection{Runtime Evaluation} 
Runtime performance is of utmost importance in autonomous driving. 
Table~\ref{tab:runtime} reports the forward pass runtime performance of  \sn in contrast to other networks. To obtain fair statistics, all measurements are repeated 10 times using all the   test data  on the same NVIDIA Tesla V100-DGXS-32GB GPU card. Obtained mean runtime values together with standard deviations are presented in Table~\ref{tab:runtime}. Our method clearly 
exhibits better performance compared to \sqa and \sqb in both projection methods, BEV and SFV. 
We observed that \un  performs slightly better than \textit{SalsaNet}. There is, however, a trade-off since \un returns relatively lower accuracies (see Table~\ref{tab:quanresults}). The reason why \un is performing faster is because of the relatively less number of kernels used in each network layer. 
We here note that the standard deviation of the \sn runtime is much less than the others. This plays a vital role in the stability of the self-driving perception modules.
Lastly, all the methods are  getting faster in case of using BEV. This can be explained by the fact that the channel number and resolution in BEV images are slightly less. 

Overall, we conclude that our network inference (a single forward pass) time can reach up to 160 Hz while providing the highest accuracies in BEV. Note that this high speed is significantly faster than the sampling rate of mainstream LiDAR scanners which typically work at  $10$Hz \cite{KittiDataset}.

%-------------------------------------------------------------------------
\section{Conclusion}
 
In this work, we presented a new deep network \sn to semantically segment  road, \ie drivable free-space, and vehicle points in real-time using 3D LiDAR data only. 
Our method differs in that \sn is input-data agnostic, that means performs equivalently well  in both BEV and SFV projections although other well-known semantic segmentation networks \cite{SqueezesegV01,SqueezesegV02,Unet} have difficulties extracting the local features in the BEV projection. 
By directly transferring image-based point-wise semantic information to the point cloud, our proposed method can automatically 
generate a large set of annotated LiDAR data required for training.

Consequently, \sn is simple, fast, and returns state-of-the-art results. Our extensive quantitative and qualitative experimental evaluations  present intuitive understanding of  the strengths and weaknesses of \textit{SalsaNet} compared to alternative methods. 
Application of \sn to bootstrap the detection and tracking processes is our planned future task in the context of autonomous driving.

%-------------------------------------------------------------------------

\bibliographystyle{IEEEtran}
% \bibliographystyle{plain}

\end{document}